\documentclass{iserc}

\usepackage{multirow}

\usepackage{graphicx}
\usepackage{caption}
\usepackage{subcaption}

\usepackage[vlined,linesnumbered,ruled]{algorithm2e}

\conference{Proceedings of the 2016 Industrial and Systems Engineering Research Conference\\
H. Yang, Z. Kong, and MD Sarder, eds.}

\title{A Probabilistic Adaptive Search System\\for Exploring the Face Space}
\author{
Andres G. Abad and Luis I. Reyes Castro\\Escuela Superior Politecnica del Litoral (ESPOL)\\Guayaquil-Ecuador\\
\vspace{0.3cm}
}
\authorlist{Abad and Reyes}
\abstractID{771}

\begin{document}

\maketitle

\begin{abstract}
Face recall is a basic human cognitive process performed routinely, e.g., when meeting someone and determining if we have met that person before. Assisting a subject during face recall by suggesting candidate faces can be challenging. One of the reasons is that the search space---the face space---is quite large and lacks structure. A commercial application of face recall is facial composite systems---such as Identikit, PhotoFIT, and CD-FIT---where a \textit{witness} searches for an image of a face that resembles his memory of a particular \textit{offender}. The inherent uncertainty and cost in the evaluation of the objective function, the large size and lack of structure of the search space, and the unavailability of the \textit{gradient} concept makes this problem inappropriate for traditional optimization methods. In this paper we propose a novel evolutionary approach for searching the face space that can be used as a facial composite system. The approach is inspired by methods of Bayesian optimization and differs from other applications in the use of the skew-normal distribution as its acquisition function. This choice of acquisition function provides greater granularity, with regularized, conservative, and realistic results.
\end{abstract}

\section*{Keywords}
Bayesian Optimization, Facial Composite Systems, Eigenfaces, Face Space, Random Face Generation

\section{Introduction}

Face recall is a basic human cognitive process performed routinely, e.g., when meeting someone and determining if we have met that person before. Assisting a subject during face recall by suggesting candidate faces can be challenging. One of the reasons is that the search space---the face space---is quite large and lacks structure.

A commercial application of face recall is facial composite systems where a \textit{witness} searches for an image of a face that resembles his memory of a particular \textit{offender}. Facial composite systems include Identikit, PhotoFIT, and CD-FIT (for a review of such systems see \cite{lindsay_handbook_2013} and the references therein.) As opposed to the way human perception works, most of these systems do not holistically consider facial features (e.g., nose, eyes, mouth) and thus present major drawbacks in their effectiveness \cite{ellis_critical_1978}. An alternative holistic approach was presented in \cite{otoole_x_1993}, where faces were reconstructed as linear combinations over a vector space. This system, unfortunately, used a large number of ``knobs'' to be adjusted manually, making its applicability difficult. Most recently, evolutionary approaches---such as genetic algorithms---have been proposed to explore the face space \cite{caldwell_tracking_1991,gibson_synthesis_2003}, giving rise to facial composite systems such as EvoFIT \cite{frowd_evofit:_2004}.


From an optimization perspective, the problem corresponds to that of finding an instance in a search space that minimizes a measure of divergence from a particular target instance. In recalling a memory, however, there is no mathematical representation of the divergence measure. Instead, the divergence function corresponds to a subjective, error-prone human oracle. The inherent uncertainty and cost in the evaluation of the objective function, the large size and lack of structure of the search space, and the unavailability of the \textit{gradient} concept makes this problem appropriate for Bayesian optimization methods.

Bayesian optimization is an optimization paradigm that regards the objective function as a probabilistic object with a prior distribution \cite{snoek_practical_2012}. Candidate instances are probabilistically generated by what is called an \textit{acquisition function} \cite{settles_active_2010} and the objective function is evaluated at each of these instances. Based on the results, the prior distribution and the acquisition function are updated and new points are generated. Bayesian optimization is usually used when evaluating the objective function is costly.

In this paper we propose a novel evolutionary approach for searching the face space that can be used as a facial composite system. The approach is inspired by methods of Bayesian optimization and differs from other applications in the use of the skew-normal distribution as its acquisition function. This choice of acquisition function provides greater granularity, with regularized, conservative, and realistic results.

\begin{figure}[h]
        \centering
        \begin{subfigure}[h]{.5\textwidth}
	\includegraphics[width=\textwidth]{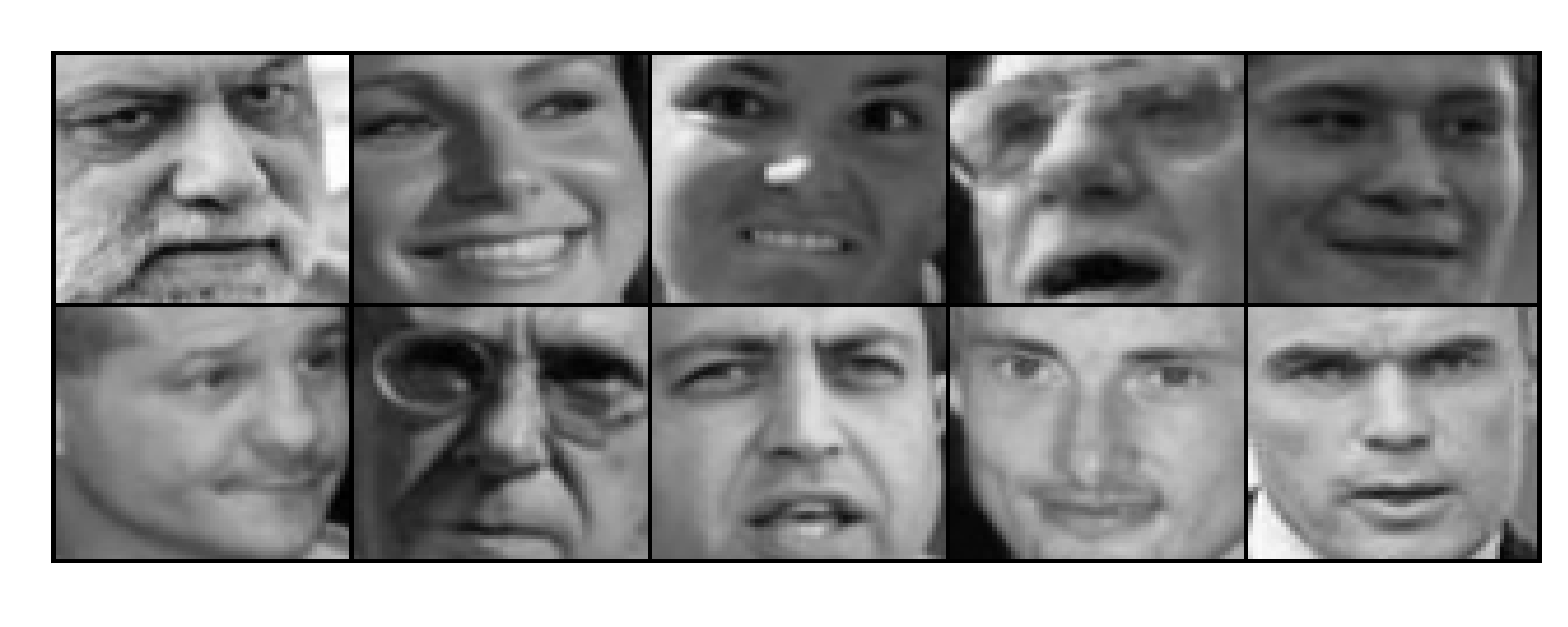}
                \caption{}
                \label{fig:faces}
        \end{subfigure}%
        ~ 
        \begin{subfigure}[h]{.5\textwidth}
	\includegraphics[width=\textwidth]{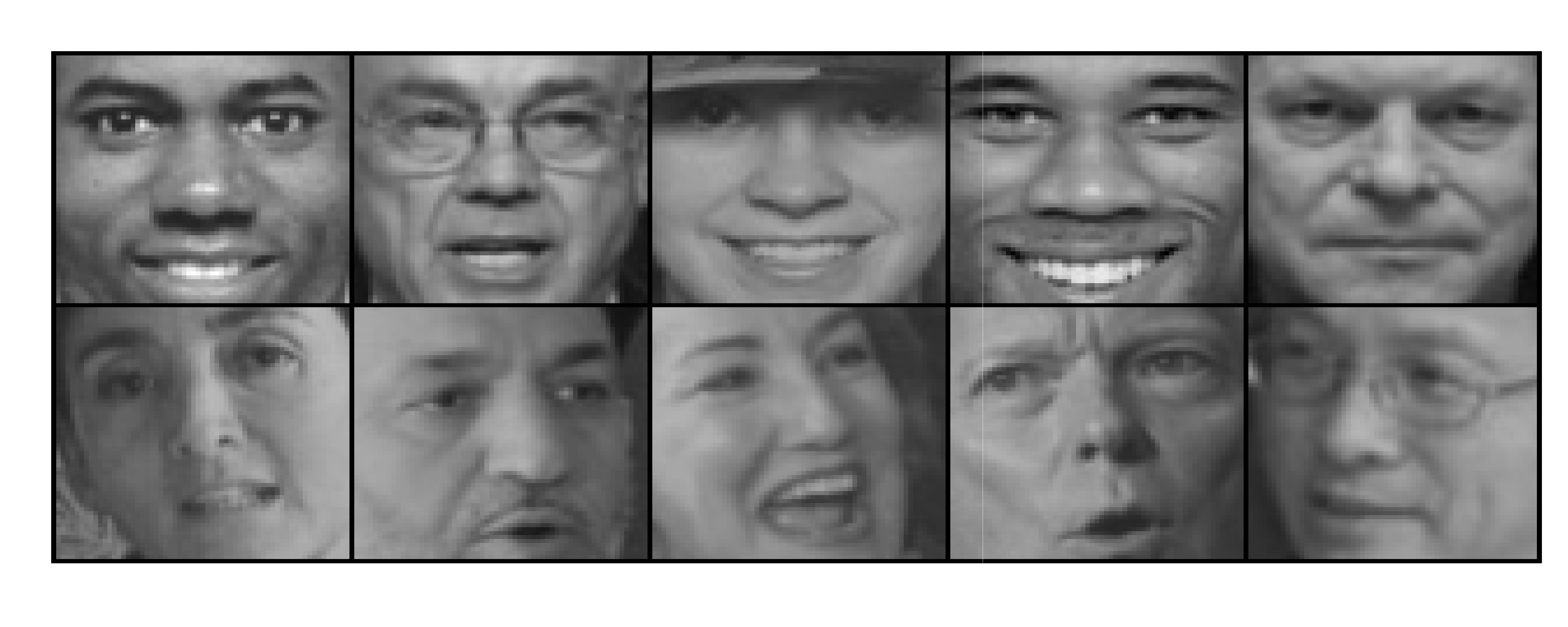}
                \caption{}
                \label{fig:symmetry}
        \end{subfigure}
             \begin{subfigure}[h]{.5\textwidth}
	\includegraphics[width=\textwidth]{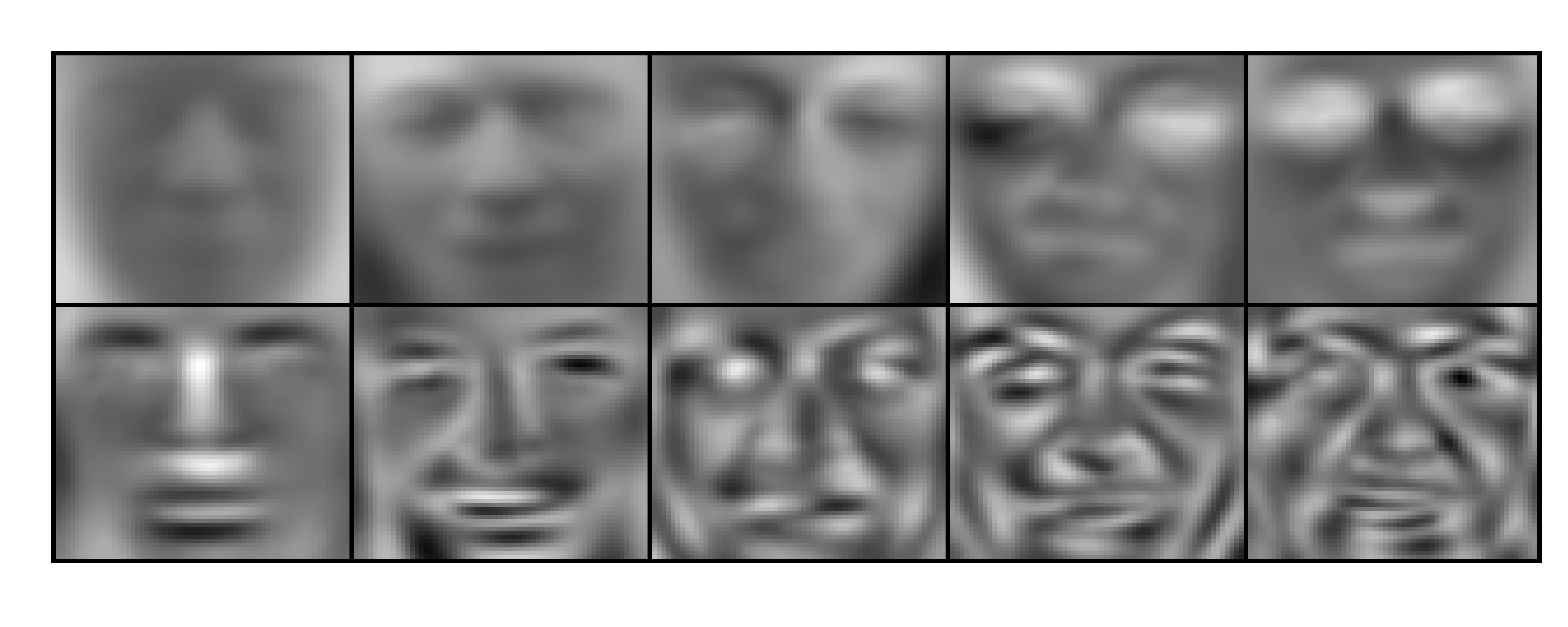}
                \caption{}
                \label{fig:eigenfaces}
        \end{subfigure}%
        ~ 
        \begin{subfigure}[h]{.5\textwidth}
	\includegraphics[width=\textwidth]{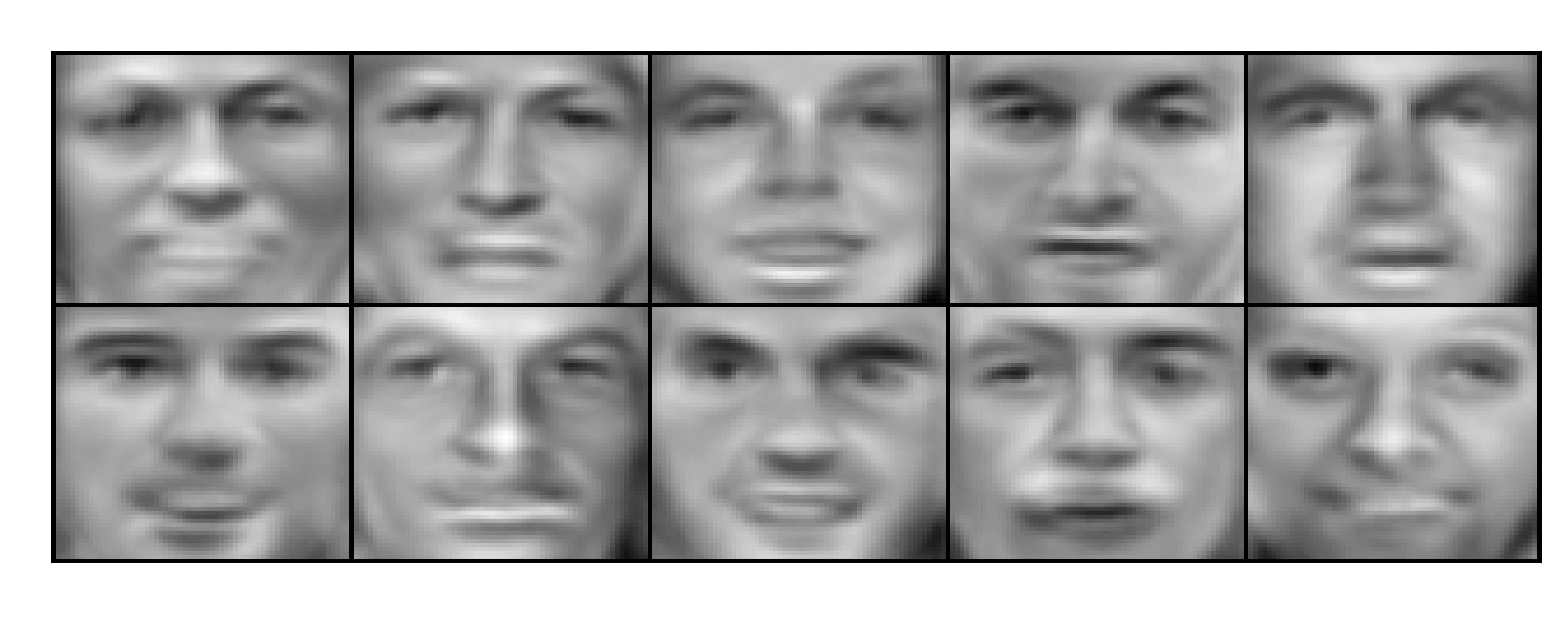}
                \caption{}
                \label{fig:random}
        \end{subfigure}
       \caption{(a) A sample from the dataset $\mathcal{X}$. (b) Faces in the top row have a high symmetry index value, while faces in the bottom row have a low value. (c) The top row shows a sample from the first ten eigenfaces ordered by the size of their associated eigenvalues, while the bottom row shows a sample of eigenfaces around the median. (d) Randomly generated faces with coordinates following a multivariate normal distribution.}\label{fig:general}
\end{figure}

We use a publicly available dataset (presented in \cite{huang_labeled_2007}) consisting of $N=13,263$ low-resolution images of faces of size $64\times64$ pixels on a grayscale. The dataset contains images of faces of world-renowned persons including politicians, artists, and sport figures. We represent the set of faces by $\mathcal{X}=\{\mathbf{x}^{(i)}\}_{i=1}^N$, where $\mathbf{x}^{(i)}\in\mathbb{R}^p$ is a vector representation of the image of a face of size $p=4,096$ ($64\times64$). Ten faces sampled from $\mathcal{X}$ are shown in Figure \ref{fig:faces}.

To improve performance, we will use exclusively ``symmetric'' faces. Here, symmetry is measured with respect to the vertical axis passing through the center of the image, measured as a quadratic difference between corresponding pixels on the left and right of the axis. Figure \ref{fig:symmetry} shows, in the top row, five faces with a high symmetry and, in the bottom row, five faces with a low symmetry. In this paper we use the top fifteen percentile symmetric faces of the original sample because improved performance was observed when using this subsample.

%

\section{Generative-Forward Model}

We propose to generate faces on a reduced representation space. The reduced representation space is obtained as the vector space spanned by the first $K$ eigenvectors of the covariance matrix of the faces in $\mathcal{X}$, known as the \textit{eigenfaces} \cite{turk_eigenfaces_1991,belhumeur_eigenfaces_1997}.

\subsection{Eigenfaces Representation}\label{sec:eigenfaces}

\begin{figure}[h]
        \centering
        \begin{subfigure}[h]{.3\textwidth}
                \includegraphics[width=\textwidth]{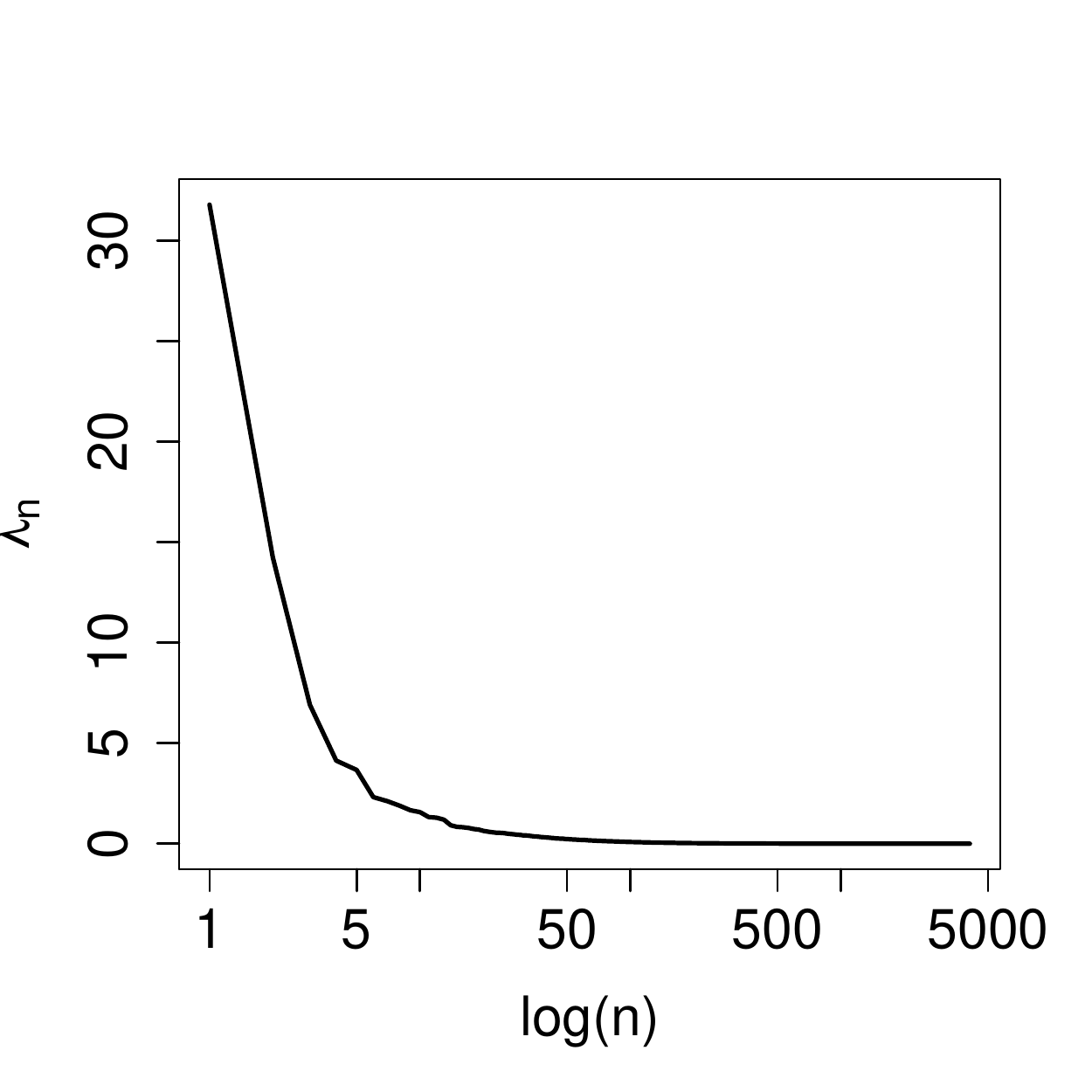}
                \caption{}
                \label{fig:eigenvalues}
        \end{subfigure}
        ~ 
        \begin{subfigure}[h]{.3\textwidth}
                \includegraphics[width=\textwidth]{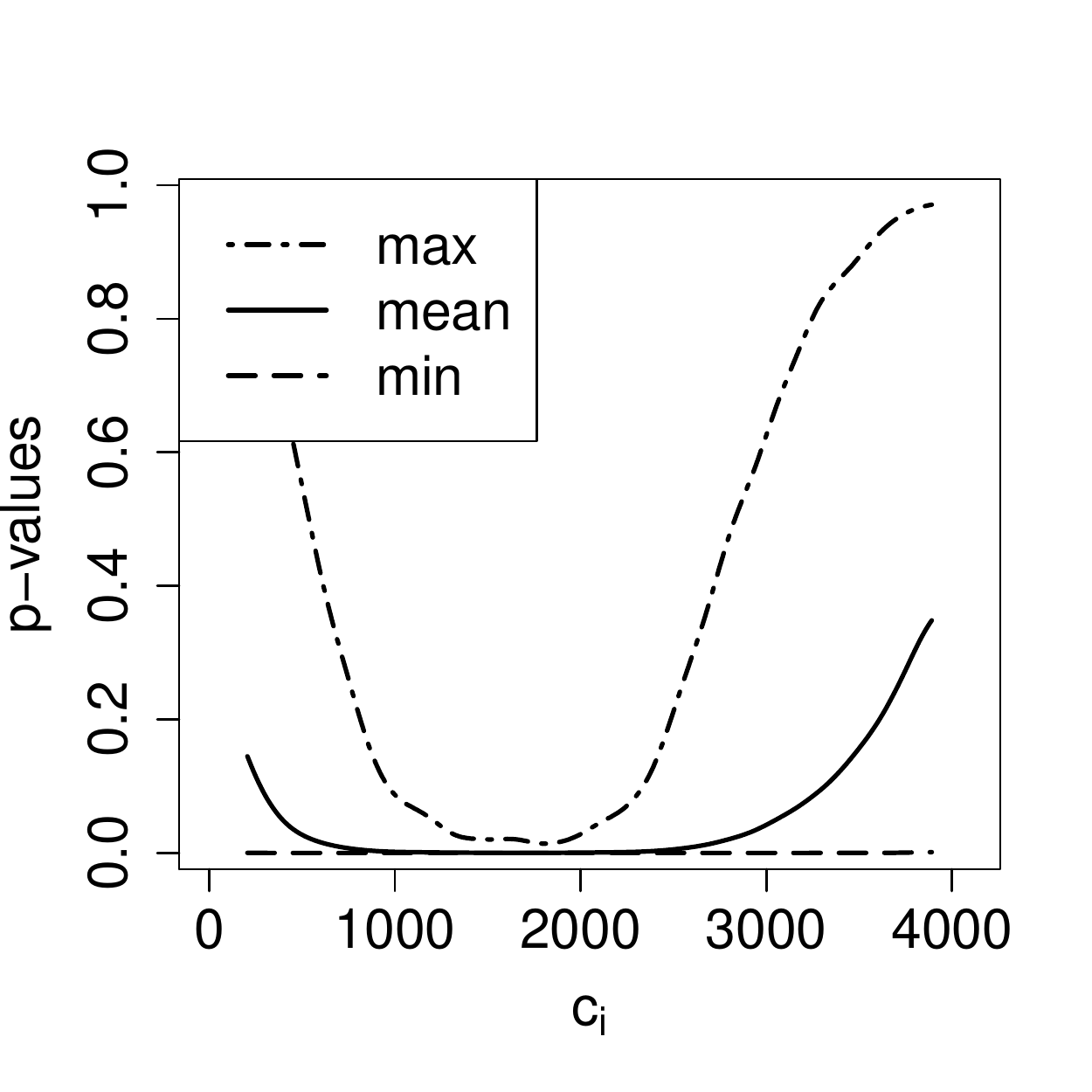}
                \caption{}
                \label{fig:pvalues}
        \end{subfigure}
         \caption{(a) Size of the eigenvalue associated with each eigenface, ordered in decreasing order. Horizontal axis is in logarithmic scale. (b) Max, mean, and Min value across $B=1000$ bootstrap replications of the $p$-value of the Shapiro-Wilks test for normality of the coordinates.}\label{fig:general}
\end{figure}

The eigenfaces ${\mathbf{v}^{(1)},\dots,\mathbf{v}^{(p)}}$ are the eigenvectors obtained from $\Sigma$, the covariance matrix of the faces in $\mathcal{X}$. Figure \ref{fig:eigenfaces} shows a sample of ten eigenfaces. Figure \ref{fig:eigenvalues} shows the amount of variability represented by each eigenface---measured by the size of their associated eigenvalue.

Let $\mathbf{c}^{(i)}\in\mathbb{R}^p$ be the coordinates of face $\mathbf{x}^{(i)}$ resulting from its projection on the space spanned by the eigenfaces ordered as the columns of matrix $\mathbf{V}=[\mathbf{v}_1|\dots|\mathbf{v}_p]$, as
\begin{equation}
\mathbf{x}^{(i)}=\mathbf{V}\mathbf{c}^{(i)}.
\end{equation}
As can be seen in Figure \ref{fig:eigenvalues}, almost all of the variability of the dataset (over $91\%$) is contained in the first $K=100$ eigenfaces. Thus, reconstructing face $\mathbf{x}^{(i)}$ using only the first $K$ eigenfaces, as 
\begin{equation}
\mathbf{x}^{(i)}\approx\mathbf{V}_{1:K}\mathbf{c}^{(i)}_{1:K},
\end{equation}
where $\mathbf{V}_{1:K}=[\mathbf{v}_1|\dots|\mathbf{v}_K]$ and $\mathbf{c}^{(i)}_{1:K}=[c^{(i)}_1,\dots,c^{(i)}_K]$, provides an efficient reduced representation of the original faces, which can be regarded as an \textit{ad-hoc} method for data compression of images of faces that takes advantage of their homogeneous structure.


\subsection{Random Faces Generation}


To generate \textit{random faces} we estimate the distribution of their coordinates on the reduced space representation. We first study the univariate distribution of elements $c_j$'s of vector $\mathbf{c}$. Figure \ref{fig:pvalues} shows the $p$-values obtained from the Shapiro-Wilk test for elements $c_j$'s, computed from a sample of size $1000$, bootstrapped $B=100$ times. The figure also shows the minimum and maximum $p$-values obtained across the $B$ bootstrapped samples. It can be seen that there is no statistical evidence to reject the hypothesis that coordinates $c_j$'s of vector $\mathbf{c}_{1:K}$ can be well represented by a Gaussian model.

Using exclusively the first $K$ eigenfaces, we run a Royston's multivariate normality test and obtained a $p$-value larger than $0.05$, thus supporting the assumption that $\mathbf{c}_{1:K}$ can be well modeled by a multivariate normal distribution as
\begin{equation}\label{eq:normal}
\mathbf{c}_{1:K}\sim\mathcal{N}_K(\mu_\mathbf{c},\Sigma_\mathbf{c}).
\end{equation}
To generate random faces we may generate multivariate random variables following a multivariate normal distribution. Figure \ref{fig:random} shows randomly generated faces obtained using the method described above and using the maximum likelihood estimators of $\mu_\mathbf{c}$ and $\Sigma_\mathbf{c}$. The \textit{smoothness} of the generated faces shown in Figure \ref{fig:random} is due to the fact that only the first $K$ eigenfaces were used, thus, resulting in compressed images.

\section{Probabilistic Adaptive Search}

Let $\tilde{\mathbf{x}}\in\mathbb{R}^K$ be the (unobserved) representation of the target face which, for our present application, is known only to the witness as a memory. If we define the search space $\mathcal{S}$ as the affine subspace given by
\begin{equation}
\mathcal{S}=\{\mu_\mathbf{c}\}\oplus\mbox{span}\{\mathbf{v}_1,\dots,\mathbf{v}_K\},
\end{equation}
we are interested in searching for faces $\mathbf{x}^*\in\mathcal{S}$ that closely resemble target face $\mathbf{\tilde{x}}$, as measured by a loss function $L:\mathbb{R}^p\times\mathbb{R}^p\rightarrow\mathbb{R}$. This corresponds to the following problem
\begin{equation}\label{eq:optimization}
\mathbf{x}^*\in\arg\min_{\mathbf{x}\in\mathcal{S}}L(\mathbf{x},\tilde{\mathbf{x}}),
\end{equation}
usually referred to as the \textit{projection problem} in the optimization literature.

Loss function $L$ is subjectively evaluated by the witness; i.e., we do not have an explicit mathematical description for it. Thus, it may be regarded as a human oracle that may only be accessed through queries. As a consequence, common optimization techniques such as gradient-based methods are not appropriate for solving problem (\ref{eq:optimization}). 

Inspired by methods of Bayesian optimization, we propose an active learning approach we termed \textit{Probabilistic Adaptive Search} to randomly generate faces that closely resemble the target face. Formally, this corresponds to the problem of finding a distribution
\begin{equation}\label{eq:optimization2}
F^*\in\arg\min_{F\in\mathcal{F}}\mathbb{E}_{\mathbf{x}\sim F}[L(\mathbf{x},\tilde{\mathbf{x}})],
\end{equation}
 where $\mathcal{F}$ is a family of parametric distributions. Note that this shifts the instances in the search space from a face $\mathbf{x}$ (in problem \ref{eq:optimization}) to a distribution of faces $F$ (in problem \ref{eq:optimization2}).
 
For the family $\mathcal{F}$ we will consider the skew-normal (SN) distribution family \cite{ohagan_bayes_1976,azzalini_class_1985}.  Skew-normal is a term used to refer to a family of parametric distributions which include the standard normal distribution as a special case and with properties mirroring the normal distribution. These properties include the fact that marginal distributions of skew-normally distributed vectors are also skew-normally distributed. Next, we provide a review of the multivariate SN distribution following the exposition in \cite{azzalini_multivariate_1996}.

\subsection{Skew-Normal Distribution Review}

Consider a random vector $\mathbf{Y}\in\mathbb{R}^K$ following a $K$-dimensional skew-normal distribution, written as
\begin{equation}
\mathbf{Y}\sim\mathcal{SN}_K(\mathbf{\lambda},\Psi),
\end{equation}
where $\mathbf{\lambda}\in\mathbb{R}^K$ and $\Psi\in\mathbb{R}^{K\times K}$ are its parameters. The mathematical expression for the SN density is
\begin{equation}
f_K(\mathbf{Y})=2\phi(\mathbf{Y};\Omega)\Phi(\mathbf{\alpha}^\intercal\mathbf{Y}),
\end{equation}
where $\phi$ is a $K$-dimensional normal density with standardized marginals and covariance matrix $\Omega$; and $\Phi$ is the one-dimensional standardized normal distribution function. The parameters of the model are defined as
\begin{equation}\alpha^\intercal=\frac{\mathbf{\lambda}^\intercal\Psi^{-1}\Delta^{-1}}{(1+\mathbf{\lambda}^\intercal\Psi^{-1}\mathbf{\lambda})^{1/2}},\end{equation}
\begin{equation}\Delta=\mbox{diag}((1-\delta_1^2)^{1/2},\dots,(1-\delta_K^2)^{1/2}),\end{equation}
\begin{equation}\mathbf{\lambda}=(\lambda(\delta_1),\dots,\lambda(\delta_K))^\intercal,\end{equation}
\begin{equation}\lambda(\delta)=\frac{\delta}{(1-\delta^2)^{1/2}},\end{equation}
\begin{equation}\Omega=\Delta(\Psi+\mathbf{\lambda}\mathbf{\lambda}^\intercal)\Delta,\end{equation}
where parameters $\lambda(\delta_i)\in(-1,1)$ regulate the degree of skewness of the distribution. In particular, $\lambda(\delta_i)=0$ corresponds to a marginal standardized normal distribution.

\section{Face Search System Description}

The proposed system for searching the face space is based on an iterative procedure, which at each iteration updates a generative model to generate faces closer to the target face. 

Specifically, the system works as follows. We begin by sampling a set of faces $\mathcal{A}^{(0)}\subset\mathcal{X}$ containing faces which the witness decides are similar to target face $\tilde{\mathbf{x}}$. At each iteration $t$ we obtain set $\mathcal{X}^{(t)}$ containing faces randomly generated following the generative model described by equations (\ref{eq:mu}) and (\ref{eq:skewmodel}). We then obtain set $\mathcal{A}^{(t)}$ by combining the elements from $\mathcal{X}^{(t)}$ which the witness decides are similar to target face $\tilde{\mathbf{x}}$, together with the elements in $\mathcal{A}^{(t-1)}$. We conclude iteration $t$ by updating probability $\mathbb{P}^{(t)}$ following equation (\ref{eq:mix}). We then iterate until the witness is satisfied with a particular generated face $\mathbf{x}^*$ resembling his memory of target face $\mathbf{\tilde{x}}$. The details of the method are presented in Algorithm \ref{alg:main}.

\begin{algorithm}
\caption{System's Algorithm}
\label{alg:main}
\KwIn{Dataset $\mathcal{X}=\{\mathbf{x}_1,\dots,\mathbf{x}_N\}$ consisting of $N$ images of faces; oracle $L(\mathbf{x}^{(i)},\mathbf{x}^{(j)})$ providing a measure of dissimilarity between $\mathbf{x}^{(i)}$ and $\mathbf{x}^{(j)}$; target face $\mathbf{\tilde{x}}$; local and global selection level $\varepsilon$ and $\varepsilon^*$, respectively}
\KwOut{A face $\mathbf{x}^*$ such that $L(\mathbf{x}^*,\tilde{\mathbf{x}})<\varepsilon^*$}
$\mathcal{A}^{(0)}=\{\mathbf{x}\in\mathcal{X}' | L(\mathbf{x},\tilde{\mathbf{x}})<\varepsilon, \mathcal{X}^{(0)}\subset\mathcal{X}\}$\\
$t=0$\\
\Repeat{$L(\mathbf{x},\mathbf{\tilde{x}})<\varepsilon^*$ for $\mathbf{x}\in\mathcal{A}^{(t)}$}{
$t=t+1$\\
Generate set $\mathcal{X}^{(t)}=\{\mathbf{x}_1^{(t)},\dots,\mathbf{x}_n^{(t)}\}$ (Equations \ref{eq:mu} and \ref{eq:skewmodel})\\
$\mathcal{A}^{(t)}=\mathcal{A}^{(t-1)}\cup\{\mathbf{x}\in\mathcal{X}^{(t)} | L(\mathbf{x},\tilde{\mathbf{x}})<\varepsilon\}$\\
Update probability distribution $\mathbb{P}^{(t)}$ (Equation \ref{eq:mix}).\\
}
\Return $\mathbf{x}^*\in\mathcal{A}^{(t)}$ such that $L(\mathbf{x}^*,\mathbf{\tilde{x}})<\varepsilon^*$
\label{system}
\end{algorithm}



We use the following \textit{Hierarchical Bayesian Model} to randomly generate faces. In our setting, parameters $\tilde{\mu}_i^{(t)}$'s are regarded as random variables with the following distribution
\begin{equation}\label{eq:mu}
\mathbf{\tilde{\mu}}^{(t)}|\mathcal{A}^{(t)}\sim\mathbb{P}^{(t)}(\mathbf{\tilde{\mu}}^{(t)}|\mathcal{A}^{(t)},\theta),
\end{equation}
where $\mathcal{A}^{(t)}=\{\mathbf{a}^{(1)},\dots,\mathbf{a}^{(N_t)}\}$ is the set of previously accepted faces and $\mathbb{P}^{(t)}$ is a mix distribution defined as
\begin{equation}\label{eq:mix}
\mathbb{P}^{(t)}(\mathbf{\tilde{\mu}}^{(t)}|\mathcal{A}^{(t)},\theta)=\frac{1}{Z}\sum_{\mathbf{a}\in\mathcal{A}^{(t)}}f(\mathbf{\tilde{\mu}}^{(t)};\mathbf{a},\theta),
\end{equation}
where $f$ is a kernel function centered at each $\mathbf{a}\in\mathcal{A}^{(t)}$, with hyperparameter vector $\theta$, and normalization factor $Z$.

Since we are restricted to $(1-\delta_i^2)\geq0$, we adjust the length of vector $\mathbf{\tilde{\mu}}^{(t)}=(\tilde{\mu}^{(t)}_1,\dots,\tilde{\mu}^{(t)}_K)$, by $k^*$ given by
\begin{equation}\label{eq:opt3}
k^*=\arg\max_k \{k | \|k\mathbf{\tilde{\mu}}^{(t)}\|_\infty\leq1-\zeta\},
\end{equation}
where a small $\zeta>0$ is used to improve numerical stability and to avoid singularity issues. Note that this adjustment preserves the direction of vector $\mathbf{\tilde{\mu}}$ and results in regularized, conservative updates. 

\begin{figure}[h]
        \centering
        \begin{subfigure}[h]{.3\textwidth}
	\centering
	\includegraphics[width=0.35\textwidth]{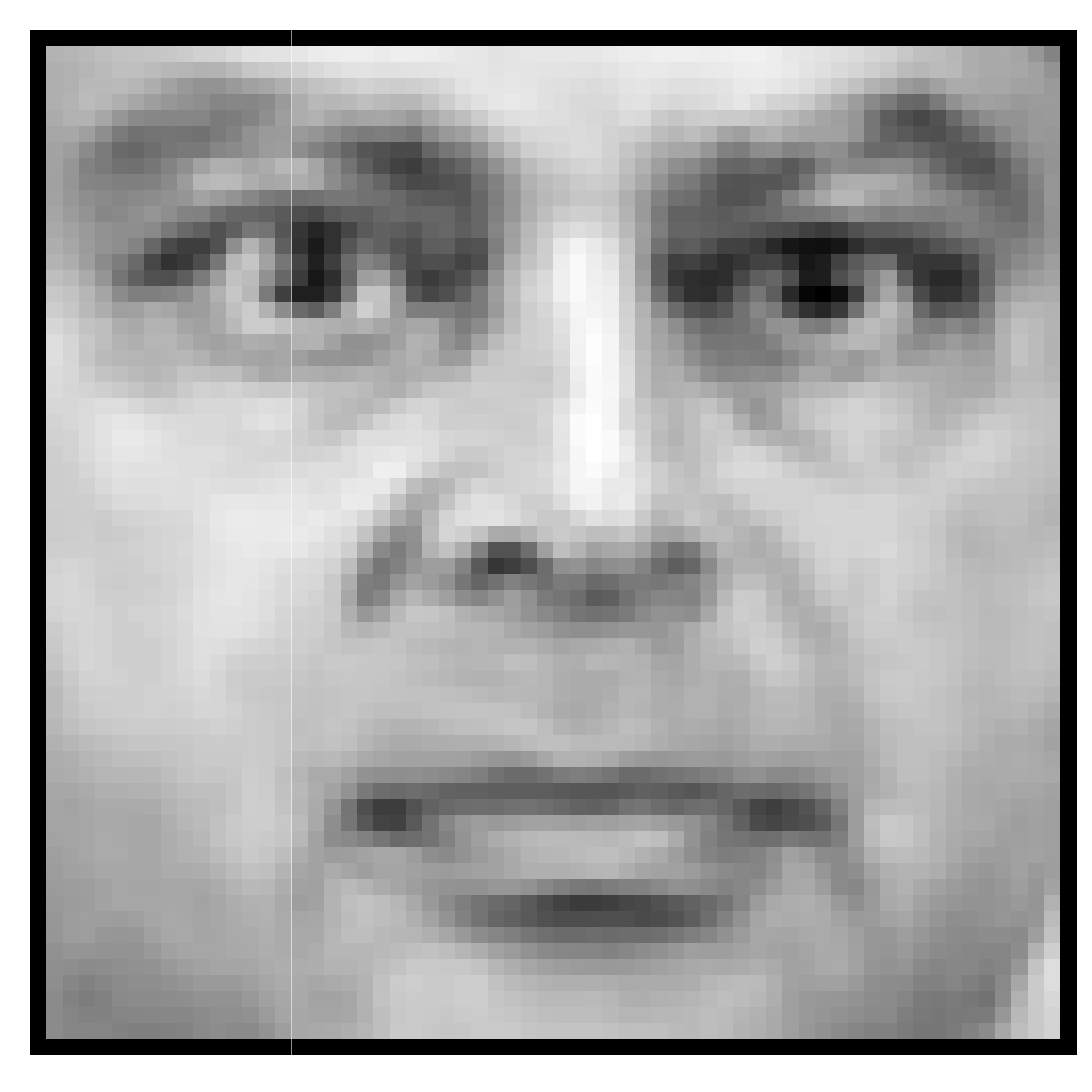}
        \end{subfigure}%
        ~ 
        \begin{subfigure}[h]{.3\textwidth}
		\includegraphics[width=\textwidth]{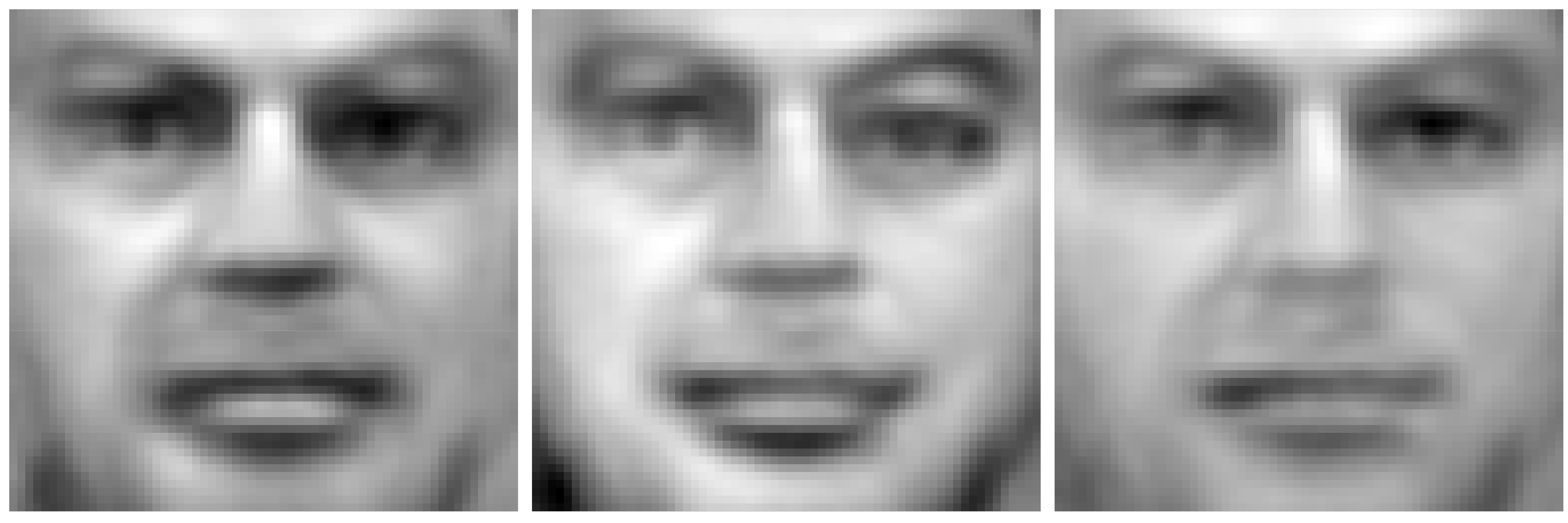}
        \end{subfigure}
        
        \begin{subfigure}[h]{.3\textwidth}
	\centering
		\includegraphics[width=0.35\textwidth]{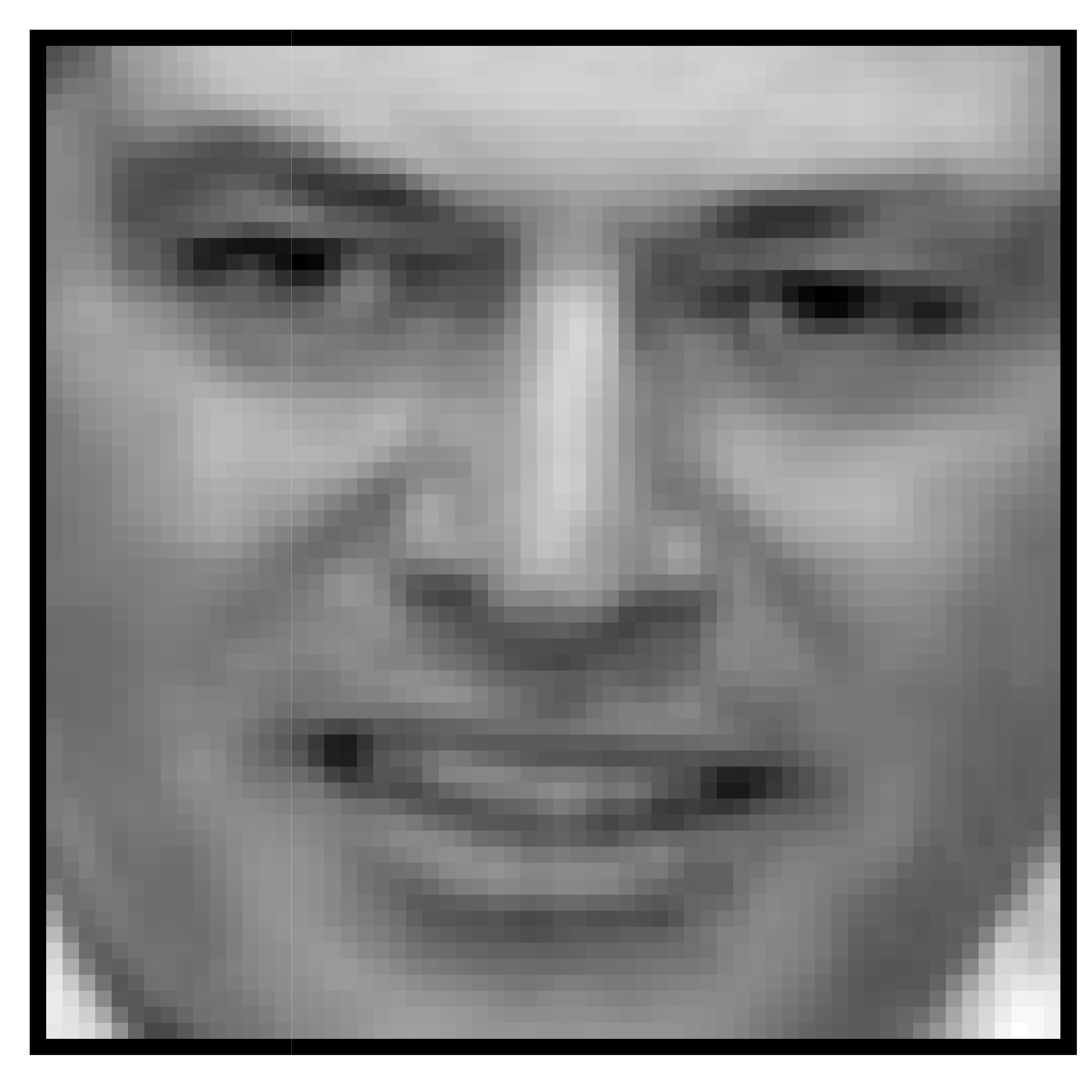}
        \end{subfigure}%
        ~ 
        \begin{subfigure}[h]{.3\textwidth}
	\includegraphics[width=\textwidth]{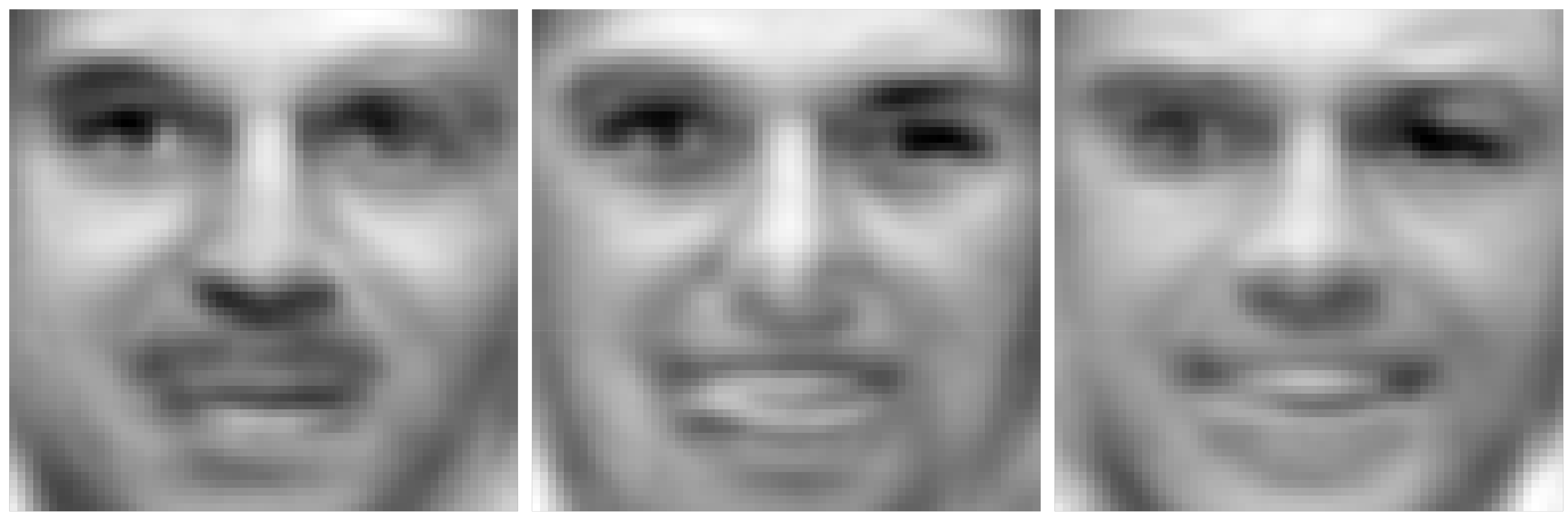}
        \end{subfigure}
        
        \begin{subfigure}[h]{.3\textwidth}
		\centering
	\includegraphics[width=0.35\textwidth]{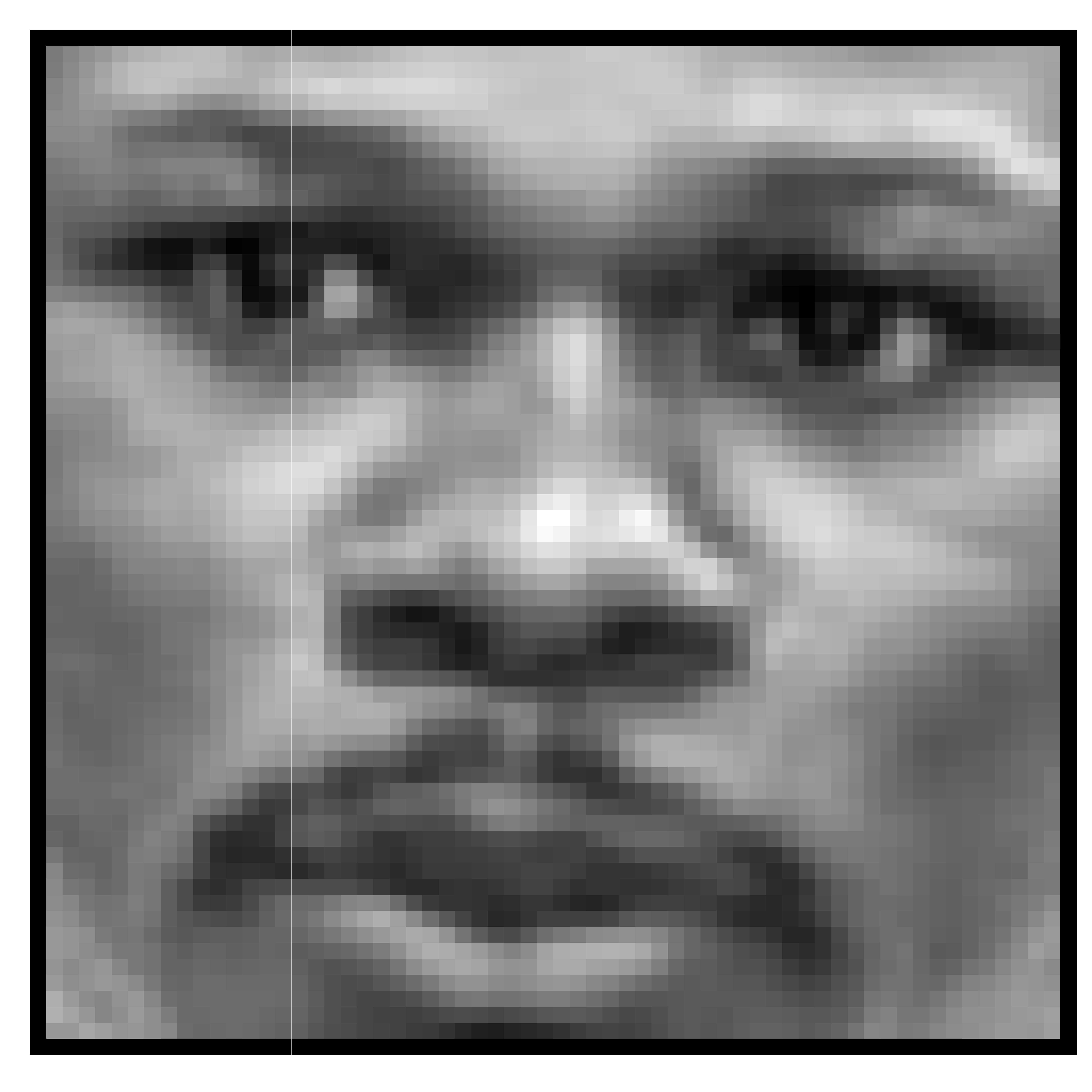}
        \end{subfigure}%
        ~ 
        \begin{subfigure}[h]{.3\textwidth}
	\includegraphics[width=\textwidth]{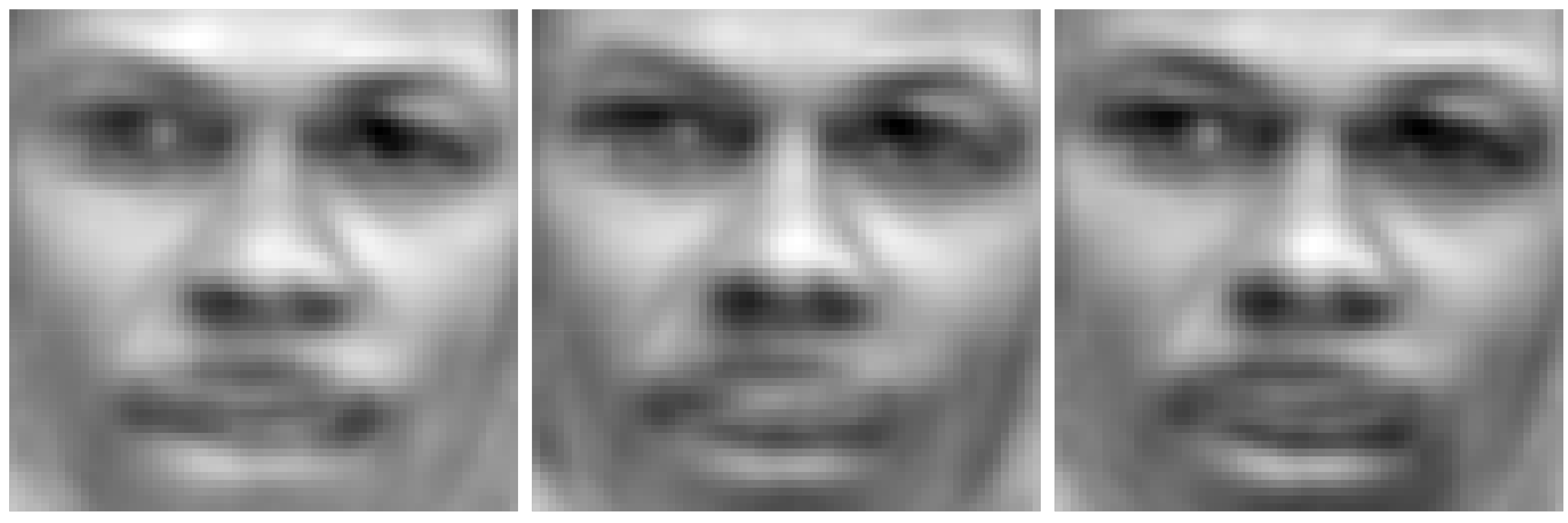}
        \end{subfigure}
       \caption{Faces on the right were generated using the skew-normal distributions, skewed towards the face $\mathbf{x}^*\in\mathcal{X}$ in the left column at each row.}\label{fig:skew}
\end{figure}

At iteration $t$, coordinates $\mathbf{c}^{(t)}$'s are generated following 
\begin{equation}\label{eq:skewmodel}
(\mathbf{c}^{(t)}-\mathbf{\mu_{\mathbf{c}}})|\mathbf{\tilde{\mu}}^{(t)}\sim\mathcal{SN}_K(\mathbf{\lambda},\Sigma_\mathbf{c}),
\end{equation}
using $\delta_i=\left(\frac{\pi}{2}\right)^{1/2}\tilde{\mu}_i^{(t)}$. Figure \ref{fig:skew} shows three faces generated from each of three target faces, using the generative model described by (\ref{eq:skewmodel}).

\section{Conclusion and Future Work}

We presented a system for exploring the face space under a Probabilistic Adaptive Search framework. The proposed system produces photo-like faces that are skewed toward a particular target face. Candidate faces are initially sampled from the original dataset, and then obtained from the updates provided by the witness. The approach uses the skew-normal distribution as an acquisition function, which provides a regularized distribution, avoiding generating unlikely instances resulting in distorted faces. The particular application considered here is that of facial composite systems. Further applications include artificially augmenting datasets with dynamic adjustments; and assisting in the design of human-perception experiments, by exploring the face space, for instance, in search of regions of interest.

\bibliography{EIGEN_FACES}

\begin{thebibliography}{10}

\bibitem{lindsay_handbook_2013}
R.~C.~L. Lindsay, David~F. Ross, J.~Don Read, and Michael~P. Toglia.
\newblock {\em The {Handbook} of {Eyewitness} {Psychology}: {Volume} {II}:
  {Memory} for {People}}.
\newblock Psychology Press, May 2013.

\bibitem{ellis_critical_1978}
Hadyn~D. Ellis, Graham~M. Davies, and John~W. Shepherd.
\newblock A critical examination of the {Photofit} system for recalling faces.
\newblock {\em Ergonomics}, 21(4):297--307, 1978.

\bibitem{otoole_x_1993}
Alice~J. O'toole and Jamie~L. Thompson.
\newblock An {X} {Windows} tool for synthesizing face images from eigenvectors.
\newblock {\em Behavior Research Methods, Instruments, \& Computers},
  25(1):41--47, March 1993.

\bibitem{caldwell_tracking_1991}
Craig Caldwell and Victor Johnston.
\newblock Tracking a {Criminal} {Suspect} through "{Face}-{Space}" with a
  {Genetic} {Algorithm}.
\newblock pages 416--421. Morgan Kaufmann Publisher, July 1991.

\bibitem{gibson_synthesis_2003}
Stuart Gibson, Alvaro~Pallares Bejarano, and Christopher Solomon.
\newblock Synthesis of photographic quality facial composites using
  evolutionary algorithms.
\newblock In {\em In {Proc}. {British} {Machine} {Vision} {Conf}. 2003}, pages
  221--230, 2003.

\bibitem{frowd_evofit:_2004}
Charlie~D. Frowd, Peter J.~B. Hancock, and Derek Carson.
\newblock {EvoFIT}: {A} {Holistic}, {Evolutionary} {Facial} {Imaging}
  {Technique} for {Creating} {Composites}.
\newblock {\em ACM Trans. Appl. Percept.}, 1(1):19--39, July 2004.

\bibitem{snoek_practical_2012}
Jasper Snoek, Hugo Larochelle, and Ryan~P Adams.
\newblock Practical {Bayesian} {Optimization} of {Machine} {Learning}
  {Algorithms}.
\newblock In F.~Pereira, C.~J.~C. Burges, L.~Bottou, and K.~Q. Weinberger,
  editors, {\em Advances in {Neural} {Information} {Processing} {Systems} 25},
  pages 2951--2959. Curran Associates, Inc., 2012.

\bibitem{settles_active_2010}
Burr Settles.
\newblock Active learning literature survey.
\newblock {\em University of Wisconsin, Madison}, 52(55-66):11, 2010.

\bibitem{huang_labeled_2007}
Gary~B. Huang, Manu Ramesh, Tamara Berg, and Erik Learned-Miller.
\newblock Labeled {Faces} in the {Wild}: {A} {Database} for {Studying} {Face}
  {Recognition} in {Unconstrained} {Environments}.
\newblock Technical Report 07-49, University of Massachusetts, Amherst, October
  2007.

\bibitem{turk_eigenfaces_1991}
Matthew Turk and Alex Pentland.
\newblock Eigenfaces for {Recognition}.
\newblock {\em Journal of Cognitive Neuroscience}, 3(1):71--86, January 1991.

\bibitem{belhumeur_eigenfaces_1997}
P.N. Belhumeur, J.P. Hespanha, and D.~Kriegman.
\newblock Eigenfaces vs. {Fisherfaces}: recognition using class specific linear
  projection.
\newblock {\em IEEE Transactions on Pattern Analysis and Machine Intelligence},
  19(7):711--720, July 1997.

\bibitem{ohagan_bayes_1976}
A.~O'hagan and Tom Leonard.
\newblock Bayes estimation subject to uncertainty about parameter constraints.
\newblock {\em Biometrika}, 63(1):201--203, January 1976.

\bibitem{azzalini_class_1985}
A.~Azzalini.
\newblock A {Class} of {Distributions} {Which} {Includes} the {Normal} {Ones}.
\newblock {\em Scandinavian Journal of Statistics}, 12(2):171--178, 1985.

\bibitem{azzalini_multivariate_1996}
A.~Azzalini and A.~Dalla Valle.
\newblock The multivariate skew-normal distribution.
\newblock {\em Biometrika}, 83(4):715--726, December 1996.

\end{thebibliography}

\end{document}